\documentclass[11pt]{article}

\usepackage[preprint]{acl}

\usepackage{times}
\usepackage{latexsym}

\usepackage[T1]{fontenc}

\usepackage[utf8]{inputenc}

\usepackage{microtype}

\usepackage{inconsolata}

\usepackage{graphicx}
\usepackage{subcaption}
\usepackage{booktabs}
\usepackage{amsmath}
\usepackage{amssymb}
\usepackage{mathtools}
\usepackage{amsthm}
\usepackage[capitalize,noabbrev]{cleveref}
\usepackage{multirow}
\usepackage{tikz}
\usepackage{pgfplots}
\usepackage{pgfplotstable}
\pgfplotsset{compat=1.18}
\usepackage{algorithm}
\usepackage{algpseudocode}
\algrenewcommand\textproc{\textnormal} 
\algrenewcommand\algorithmicfunction{\textbf{Function}} 
\newtheorem{hypothesis}{Hypothesis}

\newcommand{\rowavg}[5]{%
    \pgfmathparse{(#1 + #2 + #3 + #4 + #5) / 5}%
    \pgfmathparse{round(\pgfmathresult * 100) / 100.0}%
    \pgfmathprintnumber[fixed zerofill, precision=2]{\pgfmathresult}%
}

\newcommand{\rowavgbold}[5]{%
    \pgfmathparse{(#1 + #2 + #3 + #4 + #5) / 5}%
    \pgfmathparse{round(\pgfmathresult * 100) / 100.0}%
    \pgfmathsetmacro{\boldavg}{\pgfmathresult}%
    \pgfmathparse{int(\boldavg)}%
    \pgfmathsetmacro{\boldint}{\pgfmathresult}%
    \pgfmathparse{(\boldavg - \boldint) * 100}%
    \pgfmathroundto{\pgfmathresult}%
    \pgfmathsetmacro{\boldfrac}{int(\pgfmathresult)}%
    \textbf{\boldint.\ifnum\boldfrac<10 0\fi\boldfrac}%
}

\title{PRPO: Aligning Process Reward with Outcome Reward in Policy Optimization}

\author{
  Ruiyi Ding\textsuperscript{1},
  Yongxuan Lv\textsuperscript{2},
  Xianhui Meng\textsuperscript{2},
  Jiahe Song\textsuperscript{3},
  Chao Wang\textsuperscript{1,*},
  Chen Jiang\textsuperscript{4,5,*},
  Yuan Cheng\textsuperscript{4,5,*}\\
  \textsuperscript{1}Shanghai University, Shanghai, China \\
  \textsuperscript{2}University of Science and Technology of China, Hefei, China \\
  \textsuperscript{3}Shanghai Jiaotong University, Shanghai, China \\
  \textsuperscript{4}Artificial Intelligence Incubation and Innovation Institute, Fudan University, Shanghai, China \\
  \textsuperscript{5}Shanghai Academy of AI for Science, Shanghai, China \\
  \textsuperscript{*}Corresponding authors: cwang@shu.edu.cn, jiangchen@sais.org.cn, chengyuan@sais.org.cn 
}
\begin{document}
\maketitle

\begin{abstract}
Policy optimization for large language models often suffers from sparse reward signals in multi-step reasoning tasks. Critic-free methods like GRPO assign a single normalized outcome reward to all tokens, providing limited guidance for intermediate reasoning . While Process Reward Models (PRMs) offer dense feedback, they risk premature collapse when used alone, as early low-reward tokens can drive policies toward truncated outputs. We introduce Process Relative Policy Optimization (PRPO), which combines outcome reliability with process-level guidance in a critic-free framework. PRPO segments reasoning sequences based on semantic clues, normalizes PRM scores into token-level advantages, and aligns their distribution with outcome advantages through location-parameter shift. On MATH500, PRPO improves Qwen2.5-Math-1.5B accuracy from 61.2\% to 64.4\% over GRPO using only eight rollouts and no value network, demonstrating efficient fine-grained credit assignment within critic-free optimization.
\end{abstract}
\begin{figure*}[t]
    \centering
    \includegraphics[width=1\textwidth]{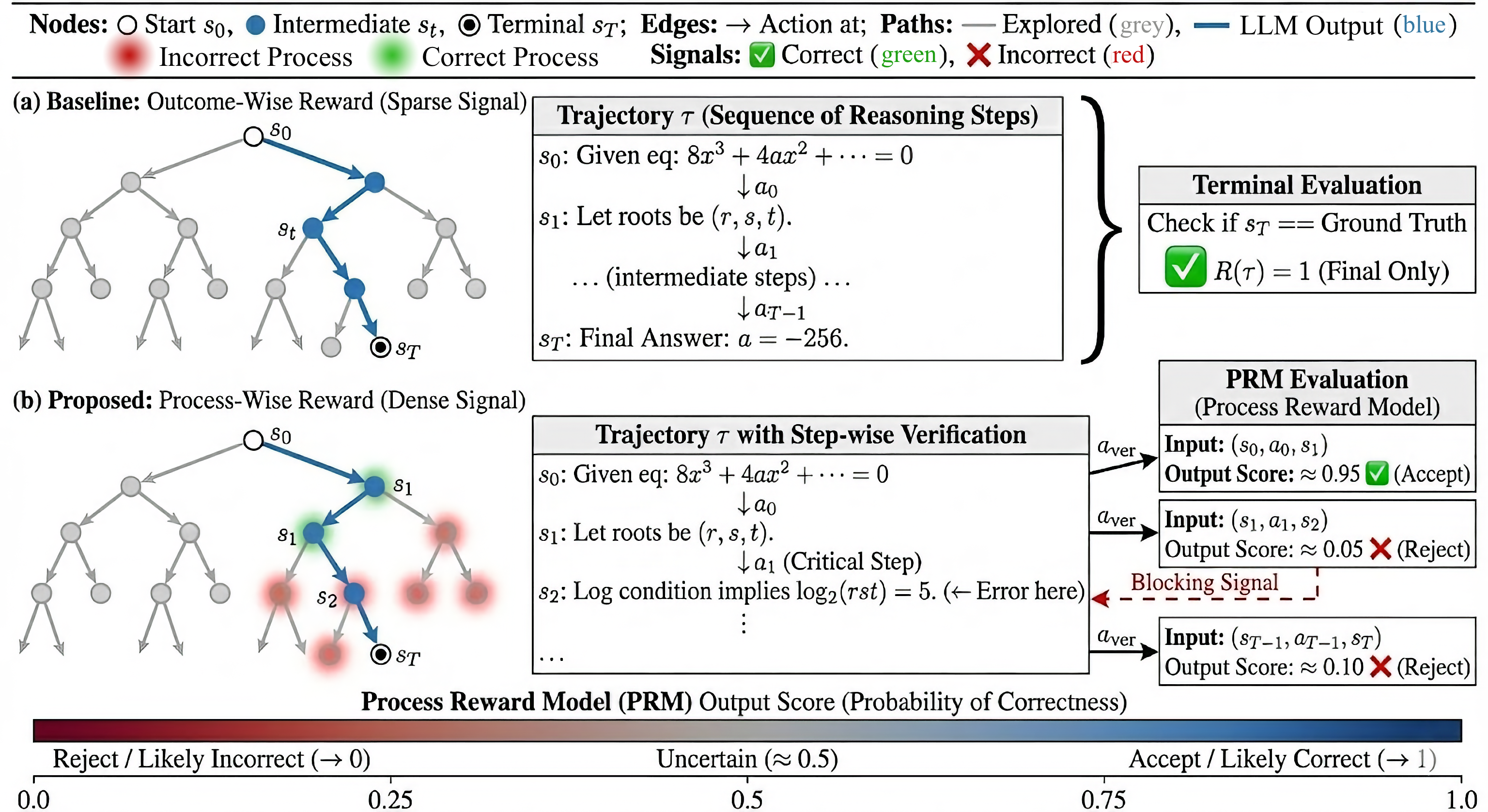}
    \caption{Comparison between outcome-level and process-level reward in RLHF.}
    \label{fig:merge-illustration}
\end{figure*}
\section{Introduction}

Reinforcement learning (RL) has become a central paradigm for aligning large language models (LLMs) with human objectives, particularly in tasks that require complex reasoning, long-horizon planning, and structured decision-making. Early RL-based alignment methods, such as reinforcement learning from human feedback (RLHF), primarily rely on sequence-level rewards that evaluate only the final generated output. While effective for short-form generation and surface-level preferences, this outcome-centric formulation exhibits fundamental limitations when applied to multi-step reasoning tasks, where correct solutions depend on a sequence of interdependent intermediate decisions rather than the final answer alone.

In long reasoning scenarios—such as mathematical problem solving, program synthesis, and multi-hop question answering—the reward signal is typically sparse, delayed, and weakly informative. A single scalar reward assigned at the end of generation provides little guidance for identifying which intermediate reasoning steps were correct, redundant, or erroneous. This exacerbates the credit assignment problem over long reasoning chains and leads to high-variance policy gradients, unstable optimization, and inefficient use of training samples. Moreover, purely outcome-based rewards fail to distinguish between reasoning trajectories that arrive at the same answer via qualitatively different processes, limiting the model's ability to internalize robust and generalizable reasoning strategies.

These challenges are further amplified in critic-free policy optimization methods, which deliberately avoid training value or critic networks in order to improve scalability and reduce computational overhead. Although such methods—exemplified by DPO-style and PPO-style algorithms—offer substantial efficiency advantages, they are inherently more sensitive to reward sparsity and noise, as they lack explicit mechanisms for temporal credit assignment. As a result, designing effective learning signals for long-horizon reasoning under critic-free constraints remains an open and critical problem.

Motivated by these limitations, recent research has progressively moved beyond sequence-level supervision toward finer-grained learning signals that explicitly model the internal reasoning process. This shift has given rise to a growing body of work on process-level and token-level preference optimization, which aims to provide denser and more informative supervision for policy learning.

In preference optimization, recent methods have progressively shifted from sequence-level to finer-grained supervision. Process-DPO \citep{lai2024step} optimizes the first erroneous reasoning process, while Full-process-DPO \citep{xu2025full} leverages self-supervised reward models to score all processes and dynamically weight gradients. TGDPO \citep{zhu2025tgdpo} introduces token-level reward guidance directly into the DPO loss, enabling dense, token-wise policy updates. These works establish a paradigm shift from sparse outcome-based rewards to dense, process-oriented supervision.

Parallel developments address sparse rewards with token-level and process-wise optimization techniques in PPO-style algorithms. TEPO \citep{lin2025token} introduces a Markov likelihood framework to link group-level rewards with token-level aggregation, reducing gradient variance in critic-free settings like GRPO \citep{shao2024grpo}. RiskPO \citep{ren2025riskpo} approaches the problem from a distributional perspective, using risk measures like Mixed Value-at-Risk to emphasize low-reward instances and prevent entropy collapse.

Despite these advances, limited research integrates Process Reward Models (PRMs) into critic-free policy optimization, and most current work didn't make use of PRM's process-level characteristics and still provide only one sparse reward signal for each output sequence. PSGPO \citep{dai2025process} explored PRM guidance within PPO, showing improvements in code generation. Other critic-free works reshape outcome rewards with PRM signals to distinguish sequences that share identical outcomes \citep{zou2025reasonflux,ren2025lsrl}, but still deliver a single reward per sequence, leaving process-level supervision underutilized. This gap motivates PRPO (Process Relative Policy Optimization), which intends to make use of the full signal from dense PRM feedback with sparse outcome rewards in a critic-free framework while retaining computational efficiency. 

It is worth noting that a current work tried to realize process-level credit assignment in critic-free framework, which is PURE Algorithm \citep{cheng2025pure}, which is a min-form credit assignment algorithm for process reward model. However, its basic idea is to choose the min value of the process reward model output, which surely can resolve possible collapse of process reward only training (which we will mention later), but will lose much information of the different process rewards along the whole output sequence, especially for the higher ones which are usually abandoned. 

PRPO addresses this critical gap by introducing a novel framework that integrates full dense PRM feedback into critic-free optimization through distribution alignment, enabling fine-grained credit assignment while preserving the computational efficiency that makes critic-free methods attractive. This innovation is particularly important because it demonstrates that stable, dense supervision can be achieved without abandoning the resource-efficient critic-free paradigm, opening new possibilities for scalable policy optimization in complex reasoning domains.

\textbf{Our contributions are threefold:} (1) a framework that integrates process supervision into critic-free policy optimization; (2) a distribution alignment technique ensuring global consistency of token-level advantages; and (3) empirical evidence that PRPO significantly improves reasoning accuracy while maintaining efficiency comparable to GRPO.

\section{Related Work}

\subsection{Critic-Free Policy Optimization}

GRPO \citep{shao2024grpo} establishes a critic-free framework by computing group-relative advantages: $\hat{A}_{\text{group}}(\tau_i) = \frac{R(\tau_i) - \text{mean}(R(\tau_i))}{\sigma(R(\tau_i)) + \epsilon}$. Building on this, ETPO \citep{wen2024entropy} decomposes credit assignment to token-level with per-token soft Bellman updates, while TEPO \citep{lin2025token} connects group rewards to tokens via Markov likelihood. RiskPO \citep{ren2025riskpo} applies Value-at-Risk theory to reshape advantages for denser signals. Test-time adaptations try to resolve OOD (Out of Distribution) Problem, featured as ETTRL \citep{liu2025ettrl}, which uses entropy-based mechanisms for diverse candidate generation to improve efficiency in original TTRL \citep{zuo2025ttrl}, and TGRPO \citep{chen2025tgrpo}, which reformats outcome rewards into process-wise signals for embodied AI.

\subsection{Process Reward Models}

Process reward models provide step-level supervision for multi-step reasoning. Early work like PRM800K \citep{song2025prmbench} relied on costly human annotation. Automated approaches subsequently emerged to reduce this cost: Math-Shepherd \citep{wang2024math} uses self-consistency sampling and symbolic verification, while Omega PRM \citep{luo2024improve} leverages MCTS-guided search to explore reasoning trajectories. Recent methods like Qwen2.5-Math-PRM \citep{zhang2025lessons} employ LLM-based judgments to generate soft labels, and EDU-PRM \citep{cao2025process} uses entropy as fork points in MCTS searches. These advances enable scalable process supervision despite potential label noise.

\section{PRPO}
\begin{figure*}[t]
    \centering
    \includegraphics[width=1\textwidth]{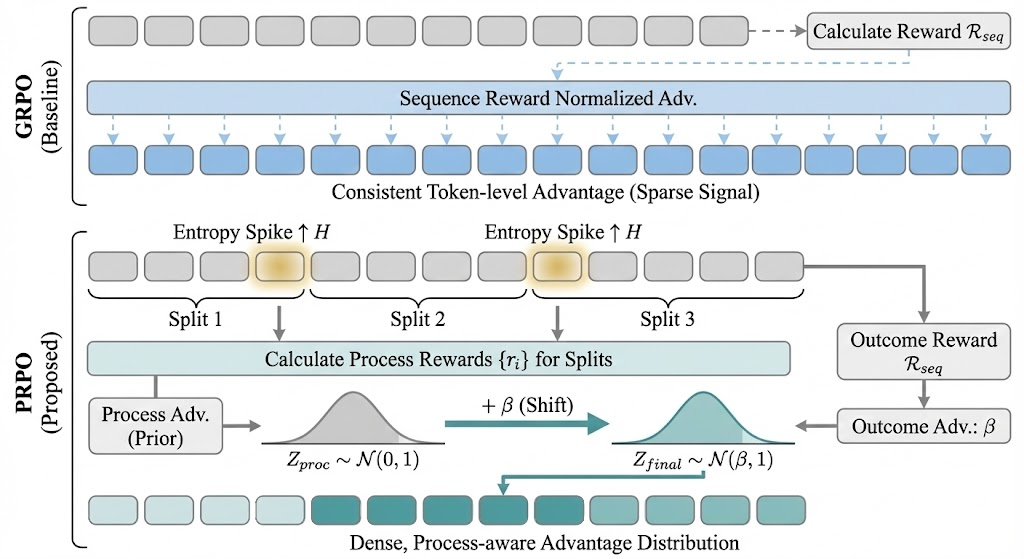}
    \caption{Comparison between GRPO and PRPO.}
    \label{fig:MethodGraph}
\end{figure*}
To leverage process-level rewards from PRMs in critic-free policy optimization, PRPO combines semantic-based segmentation with a distribution alignment mechanism that fuses outcome and process advantages. The overall framework is shown in Figure \ref{fig:MethodGraph}.

\subsection{Semantic-Based Sequence Segmentation}

Process segmentation for PRM evaluation has often relied on explicit markers such as ``Step'' \citep{zhang2025lessons}. This heuristic generalizes poorly when a model is not trained to emit such markers at every reasoning stage. To obtain a more robust and model-agnostic segmentation, we use token-level entropy to split the output sequence into segments and evaluate each segment with a PRM.

Recent studies demonstrate that token-level entropy is effective for segmenting LLM outputs \citep{liu2025ettrl, cao2025process, wang20258020}. \citet{wang20258020} find that entropy spikes frequently align with logical connectors (e.g., ``wait'', ``however'', ``nevertheless'') that bridge deductive sentences. ETTRL \citep{liu2025ettrl} similarly uses entropy spikes as fork points in rollout search, achieving better performance than traditional rollout within the TTRL framework. These results suggest that entropy peaks offer reliable semantic boundaries.

For LLM outputs, the entropy of each token is generally calculated as
\[
E_k = -\sum_{i=1}^{n} p(x_{k,i}) \log(p(x_{k,i}))
\]

Leveraging this property, we pick the top \( k \) spike points—each at least \( m \) tokens apart—as split positions and evaluate the process-level reward for every resulting segment with a PRM. 

Although it is straightforward to think entropy can decrease during training, which may make this segmentation method less reliable. However, based on empirical experiments done by \citep{wang20258020}, the tokens with higher entropy generally increase their entropy during training, while other tokens stay at a relatively stable entropy level. This means that the entropy spike points become more significant during training, thus justifying the reliability of this segmentation method. The detailed step about how this semantic-based segmentation is implemented is in Appendix \ref{app:entropy_based_segmentation}.

\subsection{Aligning Outcome-level and Process-level Reward}

Consider a sampled trajectory $\tau$ of length $T$ segmented into $M$ logical units $s_i=[t_{i-1},t_i)$ with PRM rewards $r_{\text{process}}^i$. Tokens within segment $s_i$ share the same PRM score. The segment-level process reward is expanded to tokens by setting, for any $t\in s_i$,
\[
z_{\text{process},t}(\tau) = \frac{r_{\text{process}}^i - \mu_{\text{prior\_process}}}{\sigma_{\text{prior\_process}}},
\]
giving the token vector $\mathbf{z}_{\text{process}}(\tau)\in\mathbb{R}^T$, where $\mu_{\text{prior\_process}}$ and $\sigma_{\text{prior\_process}}$ are the mean and standard deviation of the prior distribution of the PRM output. In practice, as the PRM we used (Qwen2.5-Math-PRM-7B) has its output value normalized within [0,1], we use the expected PRM output (0.5) as the mean and the standard deviation of a uniform distribution (0.289) for normalization. 

The sparse outcome reward \( R_{\text{outcome}}(\tau) \) is normalized using prior statistics to obtain the beta coefficient. As we need to fuse the outcome advantage with process advantage, we follow Dr. GRPO \citep{liu2025understand} and only use the mean value for normalization to avoid the scaling bias of the standard deviation which may cause problem in advantage aligning:
\[
\beta(\tau) = R_{\text{outcome}}(\tau) - \mu_{\text{rollout}}
\]
where \( \mu_{\text{rollout}} \) is the mean value of the outcome reward in each rollout group, which is defined as:
\[
\mu_{\text{rollout}} = \frac{1}{N} \sum_{j=1}^{N} R_{\text{outcome}}(\tau^{(j)})
\]
and broadcast this scalar to every token position of $\tau$.

\paragraph{Premature Collapse in Process-Only Reward}

Let $\pi_\theta$ be an autoregressive policy and $x_{0:T}$ a sampled sequence. Its log-likelihood decomposes as
\[
\log p_\theta(x_{0:T})=\sum_{t=0}^{T}\log \pi_\theta(x_t\mid x_{<t}).
\]
The process reward model outputs process rewards $r_t$ with standardized advantage $A_t:=r_t-\mu_{\text{process}}$. Under process-only optimization,
\begin{align}
\label{eq:process_obj}
J_{\text{proc}}(\theta)
&=\mathbb{E}_{\pi_\theta}\Big[\sum_{t=0}^{T}A_t\,\log \pi_\theta(x_t\mid x_{<t})\Big],\\
\nabla_\theta J_{\text{proc}}(\theta)
&=\mathbb{E}_{\pi_\theta}\Big[\sum_{t=0}^{T}A_t\,\nabla_\theta\log \pi_\theta(x_t\mid x_{<t})\Big].
\end{align}

Based on our observation of the process-level and outcome-level advantage during the RL training, we proposed the following hypothesis that usually causes the collapse of model trained with RL algorithm using only process-level signal:     

\begin{hypothesis}[Empirical Premature Collapse Condition during RL training with Process-Only Reward]\label{prop:early_stop}
    For a generative sequence with token-level advantages $\{A_t\}_{t=0}^T$ where $A_t = r_t - \mu_{\text{process}}$ (process reward $r_t$ relative to mean $\mu_{\text{process}}$), if there exists a position $t^\star \in [1, T]$ such that:
    \begin{enumerate}
        \item The parameters $a$ and $b$ are sampled from the same sequence: $a = \left| \frac{1}{t^\star} \sum_{i=0}^{t^\star-1} A_i \right|$ (absolute mean advantage over prefix $[0, t^\star)$) and $b = A_{t^\star}$ (advantage at position $t^\star$), where $a$ reflects early tokens with process rewards below the mean (negative advantage) and $b$ reflects a token with process reward above the mean (positive advantage).
        \item Early tokens have negative average advantage: $\frac{1}{t^\star} \sum_{i=0}^{t^\star-1} A_i = -a < 0$ with $a > 0$, indicating process rewards below the mean.
        \item Token at $t^\star$ has positive advantage: $A_{t^\star} = b > 0$, indicating process reward above the mean.
        \item The cumulative negative magnitude exceeds the positive benefit: $a \cdot t^\star > b$.
    \end{enumerate}
    Then the conflicting gradient signals from early negative advantages ($-a$) and later positive advantage ($b$) within the same sequence cause $p_\theta(x_{0:t^\star})$ to decrease, leading to premature collapse.
\end{hypothesis}
The mathematical reason for this hypothesis is shown in Appendix \ref{app:early_stop_proof}

\textbf{Interpretation} Negative advantages hit every prefix term multiplicatively, while the single positive advantage at $t^\star$ cannot compensate for the accumulated decay. Heavy-tailed or outlier negative rewards (large $a$) accelerate the collapse.

Based on this hypothesis, it can be witnessed that the root problem of the collapse of model trained with process-only reward is the fluctuation of process rewards, which can cause conflict of advantage signal. PURE \citep{cheng2025pure} resolves this problem by simply choosing the min value of the process reward model output, which surely can resolve possible collapse of process reward only training, but will lose much information of the different process rewards along the whole output sequence, especially for the higher ones which are usually abandoned. 

\textbf{Practical observation} We actually did an experiment on Qwen2.5-Math-7B model trained with process advantages alone, and found that it empirically shortened responses and eventually produced meaningless short strings, matching the hypothesis. 

\textbf{Implication} Even perfectly estimated process rewards are unsafe when used alone with relative advantages. Necessary but relatively poor processes get suppressed, and locally good-but-misaligned processes get over-weighted. A prior or alignment term (next section) is needed to prevent collapse. The prior keeps high-prior processes from being suppressed, while distribution alignment ties process rewards to outcome rewards so the final outcome advantage safely guides process-level updates.

\paragraph{Distribution Alignment}

PRPO shifts the process-advantage distribution using the outcome advantage, as shown in the bottom of Figure \ref{fig:MethodGraph}:
\[
\mathbb{E}[\mathcal{P}_{\text{process}}'] = \mathbb{E}[\mathcal{P}_{\text{process}}] +\mathbb{E}[\mathcal{P}_{\text{outcome}}],
\]
producing a fused per-token advantage
\[
\mathbf{AF}(\tau) = \mathbf{z}_{\text{process}}(\tau) + \beta(\tau),
\]
Concretely, $AF_t(\tau)=z_{\text{process},t}(\tau)+\beta(\tau)$ for every token position $t$, so each token inherits the segment-normalized PRM score plus a broadcasted, group-centered outcome shift. The policy loss becomes
\[
\mathcal{L}_{\text{PRPO}} = \mathbb{E}_{\tau_i} \left[ \mathbf{AF}(\tau)(x_t^{(i)}) \log \frac{\pi_\theta(x_t^{(i)} | x_{<t}^{(i)})}{\pi_{\text{ref}}(x_t^{(i)} | x_{<t}^{(i)})} \right].
\]

\section{Experiments}
\begin{table}[H]\footnotesize
    \centering
    \footnotesize
    \caption{Time cost analysis when one training use the PRM server and three training use the server simultaneously.}
    \label{tab:time_cost_analysis}
    \begin{tabular}{@{}lcc@{}}
    \toprule
    Method & \multicolumn{2}{c}{Time Cost (s)} \\
    \midrule
    w/o PRM step time & 46.8 \\
    w PRM step time (1 training) & 80.8 \\
    w PRM step time (3 trainings) & 84.0 \\
    \bottomrule
    \end{tabular}
\end{table}
\begin{figure*}[t]
    \centering
    \begin{subfigure}[t]{0.48\textwidth}
        \centering
        \begin{tikzpicture}
            \begin{axis}[
                width=\linewidth,
                height=5cm,
                xlabel={Epoch},
                ylabel={Accuracy},
                legend style={at={(0.03,0.97)},anchor=north west, font=\scriptsize},
                grid=both,
                grid style={dashed, gray!30},
                ymin=0.55,
                ymax=0.7,
                xmin=-1,
                xmax=11,
            ]
            \addplot[
                color=blue,
                mark=*,
            ]
            table [skip first n=1, x expr=\coordindex, y index=0, col sep=comma] {data/trainEpoch.csv};
            \addlegendentry{GRPO}
            \addplot[
                color=orange,
                mark=triangle*,
            ]
            table [skip first n=1, x expr=\coordindex, y index=4, col sep=comma] {data/trainEpoch.csv};
            \addlegendentry{PURE}
            \addplot[
                color=green!70!black,
                mark=triangle*,
            ]
            table [skip first n=1, x expr=\coordindex, y index=2, col sep=comma] {data/trainEpoch.csv};
            \addlegendentry{GRPO+PRPO}
            \end{axis}
        \end{tikzpicture}
        \subcaption{GRPO vs GRPO+PRPO.}
        \label{fig:lineplot-trainepoch-GRPO}
    \end{subfigure}
    \hfill
    \begin{subfigure}[t]{0.48\textwidth}
        \centering
        \begin{tikzpicture}
            \begin{axis}[
                width=\linewidth,
                height=5cm,
                xlabel={Epoch},
                ylabel={Accuracy},
                legend style={at={(0.03,0.97)},anchor=north west, font=\scriptsize},
                grid=both,
                grid style={dashed, gray!30},
                ymin=0.55,
                ymax=0.7,
                xmin=-1,
                xmax=11,
            ]
            \addplot[
                color=orange,
                mark=triangle*,
            ]
            table [skip first n=1, x expr=\coordindex, y index=4, col sep=comma] {data/trainEpoch.csv};
            \addlegendentry{PURE}
            \addplot[
                color=red,
                mark=square*,
            ]
            table [skip first n=1, x expr=\coordindex, y index=1, col sep=comma] {data/trainEpoch.csv};
            \addlegendentry{PRM-Avg}

            \addplot[
                color=purple,
                mark=diamond*,
            ]
            table [skip first n=1, x expr=\coordindex, y index=3, col sep=comma] {data/trainEpoch.csv};
            \addlegendentry{PRM-Avg+PRPO}
            \end{axis}
        \end{tikzpicture}
        \subcaption{PRM-Avg vs PRM-Avg+PRPO.}
        \label{fig:lineplot-trainepoch-PRM-Avg}
    \end{subfigure}
    \caption{Training accuracy on MATH. PRPO steadily improves both baselines.}
    \label{fig:train-curves}
\end{figure*}
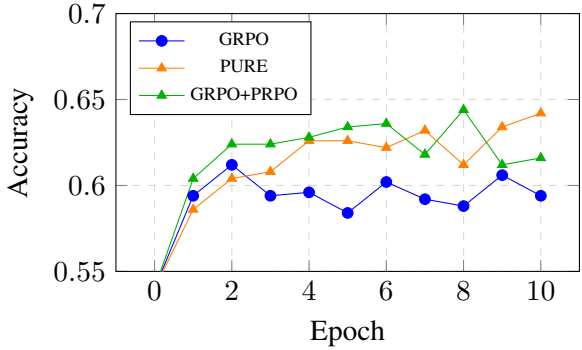
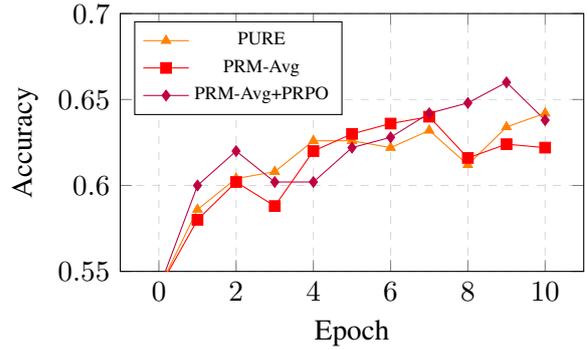
\pgfplotstableread[col sep=comma]{data/main_experiment_result.csv}\MainResultData
\begin{table*}[t]
\centering
\footnotesize
\caption{Accuracy comparison across datasets (\%).}
\label{tab:main_experiment_result}
\pgfplotstabletypeset[
    col sep=comma,
    string type,
    columns={Model,Algorithm,MATH,AMC2023,AIME2025,AIME2024},
    columns/Model/.style={
        column name={Model},
        column type=c,
        postproc cell content/.code={%
            \pgfmathtruncatemacro{\row}{\pgfplotstablerow}%
            \ifnum\row=0\relax
                \pgfkeyssetvalue{/pgfplots/table/@cell content}{\multirow{6}{*}{\rotatebox{90}{Qwen-M-1.5}}}%
            \else\ifnum\row=6\relax
                \pgfkeyssetvalue{/pgfplots/table/@cell content}{\multirow{6}{*}{\rotatebox{90}{Qwen-M-7}}}%
            \else
                \pgfkeyssetvalue{/pgfplots/table/@cell content}{}%
            \fi\fi
        }
    },
    columns/Algorithm/.style={column type=l},
    columns/MATH/.style={column name={\shortstack{MATH\\pass@1}}},
    columns/AMC2023/.style={column name={\shortstack{AMC2023\\mean32}}},
    columns/AIME2025/.style={column name={\shortstack{AIME2025\\mean32}}},
    columns/AIME2024/.style={column name={\shortstack{AIME2024\\mean32}}},
    every head row/.style={before row=\toprule, after row=\midrule},
    every row no 5/.style={after row=\midrule},
    every last row/.style={after row=\bottomrule}
]{\MainResultData}
\end{table*}

\begin{table*}[t]\footnotesize
\centering
\footnotesize
\caption{Pass rate comparison on Qwen2.5-Math-7B.}
\label{tab:math_performance}
    \begin{tabular}{l c c c c c c}
        \toprule
        \multirow{2}{*}{Method} & \multicolumn{5}{c}{Pass Rate (\%)} & \multirow{2}{*}{Average} \\
        \cmidrule(lr){2-6}
        & pass@8 & pass@16 & pass@32 & pass@64 & pass@128 & \\
        \midrule
        \multicolumn{7}{c}{\textbf{AIME 2024}} \\
        GRPO & 36.67 & 50.00 & 60.00 & 56.67 & 63.33 & \rowavg{36.67}{50.00}{60.00}{56.67}{63.33} \\
        PRMAVG & 26.67 & 33.33 & 33.33 & 43.33 & 53.33 & \rowavg{26.67}{33.33}{33.33}{43.33}{53.33} \\
        PURE & \underline{50.00} & 50.00 & \underline{63.33} & 60.00 & 63.33 & \rowavg{50.00}{50.00}{63.33}{60.00}{63.33} \\
        GRPO+PRPO & 46.67 & \underline{53.33} & 60.00 & \underline{63.33} & 60.00 & \rowavg{46.67}{53.33}{60.00}{63.33}{60.00} \\
        PRMAVG+PRPO & 46.67 & \underline{53.33} & 60.00 & 63.33 & \underline{70.00} & \rowavgbold{46.67}{53.33}{60.00}{63.33}{70.00} \\
        \midrule
        \multicolumn{7}{c}{\textbf{AIME 2025}} \\
        GRPO & 24.14 & 27.59 & \underline{37.93} & 27.59 & 41.38 & \rowavg{24.14}{27.59}{37.93}{27.59}{41.38} \\
        PRMAVG & 20.69 & 24.14 & 27.59 & 34.48 & 37.93 & \rowavg{20.69}{24.14}{27.59}{34.48}{37.93} \\
        PURE & 27.5 & 31.03 & 34.48 & 37.93 & 44.83 & \rowavg{27.5}{31.03}{34.48}{37.93}{44.83} \\
        GRPO+PRPO & \underline{31.03} & 31.03 & 31.03 & 37.93 & \underline{48.28} & \rowavgbold{31.03}{31.03}{31.03}{37.93}{48.28} \\
        PRMAVG+PRPO & 20.69 & \underline{34.48} & 34.48 & \underline{41.38} & 41.38 & \rowavg{20.69}{34.48}{34.48}{41.38}{41.38} \\
        \midrule
        \multicolumn{7}{c}{\textbf{AMC 2023}} \\
        GRPO & 77.50 & 77.50 & 82.50 & 90.00 & 92.50 & \rowavg{77.50}{77.50}{82.50}{90.00}{92.50} \\
        PRMAVG & 80.00 & 80.00 & 85.00 & 87.50 & 87.50 & \rowavg{80.00}{80.00}{85.00}{87.50}{87.50} \\
        PURE & \underline{90.00} & 87.50 & 90.00 & 95.00 & 97.50 & \rowavg{90.00}{87.50}{90.00}{95.00}{97.50} \\
        GRPO+PRPO & 85.00 & 85.00 & 90.00 & \underline{97.50} & 97.50 & \rowavg{85.00}{85.00}{90.00}{97.50}{97.50} \\
        PRMAVG+PRPO & 85.00 & \underline{90.00} & \underline{95.00} & \underline{97.50} & \underline{100.00} & \rowavgbold{85.00}{90.00}{95.00}{97.50}{100.00} \\
        \bottomrule
    \end{tabular}
\end{table*}
\begin{table*}[t]\footnotesize
    \centering
    \footnotesize
    \caption{Ablation study (\%).}
    \label{tab:ablation_study_result}
    \begin{tabular}{@{}lcc@{}}
    \toprule
    Model & Algorithm & MATH pass@1 \\
    \midrule
    \multirow{3}{*}{Qwen2.5-Math-1.5B} 
    & GRPO+\textbf{PRPO} w Random Split & 2.4 \\
    & GRPO+\textbf{PRPO} w Uniform Split & 29.8 \\
    & GRPO+\textbf{PRPO} w Entropy Based split & \textbf{64.4} \\
    \midrule
    \multirow{2}{*}{Qwen2.5-Math-1.5B} 
    & GRPO+\textbf{PRPO} w Relative distribution for process reward & 64.0 \\
    & GRPO+\textbf{PRPO} w Predefined distribution for process reward & \textbf{64.4} \\
    \bottomrule
\end{tabular}
\end{table*}

We train on 8 H200 GPUs with batch size 128, learning rate $1 \times 10^{-6}$, KL loss coefficient 0.001, clip ratio 0.2, maximum new tokens 2048, and rollout number 8. For segmentation, we pick five entropy spikes that are at least ten tokens apart ($k{=}5, n{=}10$).

For PRM, we choose Qwen2.5-Math-PRM-7B as the PRM model, and deploy it on 8 H200 GPUs according to official guidance to normalize the output value with $[0,1]$ range. We did an analysis on the time cost of PRM when training Qwen-Math-1.5B in Table \ref{tab:time_cost_analysis}. In this experiment, we take GRPO as model training without PRM, and PRPO+GRPO as model training with PRM. We trained each model for 1 epoch and calculated the average time cost for each step. Even with PRM and training three models simultaneously, the additional time cost is about 37 seconds each step (46.8 s vs 84.0 s). Comparing to the benefit and higher upper bound brought by PRM, the time cost increase is acceptable.

We compare PRPO with GRPO and PRM-Avg \citep{zou2025reasonflux,ren2025lsrl}. PRM-Avg adds the average process reward to the outcome reward, which is a common practice in recent work of using PRM in RL training like ReasonFlux \citep{zou2025reasonflux} and lsrl \citep{ren2025lsrl}. It's worth noting that there's a slight difference between lsrl and reasonflux, where lsrl use a discounted sum of process reward, while reasonflux use naive average of process reward. As reasonflux is a more recent and widely referenced work, we use naive average used in reasonflux for PRM-avg. As PRPO provides another insight of advantage calculation, it can apply to both baselines. We also compare our algorithm with the current state-of-the-art baseline PURE \citep{cheng2025pure} algorithm, which is the only critic-free algorithm that tried to realize process-level credit assignment in critic-free framework. We implement PURE with the exact same reward and advantage settings in the original work to reach the optimal performance, where the output reward signal is 0-1 reward, and the assignment temperature is set as 0.1. We evaluate performance of each algorithm on Qwen2.5-Math-1.5B and Qwen2.5-Math-7B \citep{yang2024qwen25math}. 

We conduct our algorithm and train the models with the VeRL training framework \citep{sheng2025verl}. 

Models are trained on the training split of MATH \citep{hendrycksmath2021} with 12000 samples with early stopping. Evaluation covers MATH, AMC 2023, AIME 2025, and AIME 2024 \citep{aime_official_website,amc_official_website} using greedy decoding (pass@1) for MATH as there is sufficient number of test cases in its test split (500 samples) and sampling 32 rollouts for AMC/AIME with temperature 0.7, top-p 0.9, and seeds fixed to 42 as there are limited test cases in AIME (30 samples each dataset) and AMC 2023 (40 samples). For algorithms other than PURE, rule based outcome rewards are $\pm 1$ based on SymPy-verified correctness, plus a length penalty beyond 1024 tokens. Process rewards come from Qwen2.5-Math-PRM-7B.

We draw the accuracy curve of different algorithms trained on Qwen2.5-Math-1.5B in Figure \ref{fig:train-curves}. As shown in the figure, PRPO steadily improves both baselines, and generally fit quicker and reach higher accuracy than all baselines. 

The length penalty is described as follows:
\[
\text{length\_penalty} = \begin{cases}
0 & \text{if } \text{length} \leq 1024 \\
\frac{\text{length}}{1024} & \text{if } \text{length} > 1024
\end{cases}
\]

As shown in Table \ref{tab:main_experiment_result}, our PRPO algorithm consistently outperforms both GRPO and PRM-Avg baselines across all evaluated datasets and model sizes. Qwen-M-1.5 refers to Qwen2.5-Math-1.5B model, and Qwen-M-7 refers to Qwen2.5-Math-7B model. For Qwen2.5-Math-1.5B, PRPO improves MATH pass@1 accuracy from 61.20\% (GRPO) to 64.4\% (+3.2\%), and further to 66.00\% when combined with PRM-Avg (64.00\% vs 66.00\%). On AMC2023 and AIME2024, our method also yields substantial gains, demonstrating its effectiveness in both standard and challenging mathematical benchmarks. Although our method does not perform as well as PRM-Avg on AIME2025, the gap is not significant (5.9\% vs 7.00\%), which can be attributed to capacity and fitting limits under dense supervision. PRM-Avg injects a dense per-segment signal; smaller or weaker models may not have enough capacity to disentangle noisy PRM scores from outcome supervision, effectively turning dense rewards into a length or variance penalty. This is especially pronounced on AIME2024 with the 7B model and on AIME2025 with the 1.5B model, where the tasks are harder and label noise is higher. In contrast, larger models (7B on AIME2025) can better absorb and align dense signals, which explains why the degradation diminishes as capacity increases. Comparing to PURE \citep{cheng2025pure}, our algorithm also demonstrates its robustness across all datasets. 

We also did an experiment on different pass rate accuracy of Qwen2.5-Math-7B model trained with different methods. As shown in Table \ref{tab:math_performance}, our algorithm consistently outperforms both GRPO and PRM-Avg baselines across all evaluated datasets, which shows the stability and effectiveness of our algorithm when applied to models with enough parameters. It is worth noting that PRMAVG usually underperforms GRPO on AIME2025 and AIME2024, which shows the lack of robustness of using average process reward as outcome reward. However, with PRPO, both PRMAVG and GRPO can generally obtain great improvement on all datasets under different pass rate, showing the robustness of our algorithm.

These results validate our hypothesis that aligning process-level and outcome-level rewards enables more stable and effective policy optimization, especially in multi-process reasoning tasks where sparse signals are insufficient for robust learning.

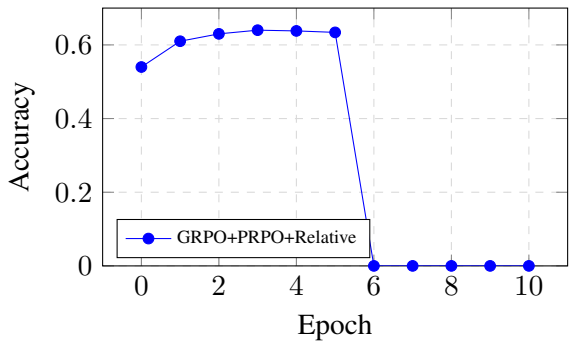
\begin{figure}[t]
    \centering
    \begin{tikzpicture}
        \begin{axis}[
            width=\linewidth,
            height=5cm,
            xlabel={Epoch},
            ylabel={Accuracy},
            legend style={at={(0.03,0.03)},anchor=south west, font=\scriptsize},
            grid=both,
            grid style={dashed, gray!30},
            ymin=0,
            ymax=0.7,
            xmin=-1,
            xmax=11,
        ]
        \addplot[
            color=blue,
            mark=*,
        ]
        table [skip first n=1, x expr=\coordindex, y index=1, col sep=comma] {data/comp.csv};
        \addlegendentry{GRPO+PRPO+Relative}
        \end{axis}
    \end{tikzpicture}
    \caption{Ablation on relative process-reward normalization.}
    \label{fig:ablation_study_result}
  \end{figure}

Additionally, we did an ablation study on our entropy based split strategy. For comparison, we select random k points that are at least n tokens away from other points as random split, and equally divide the output sequence into k+1 parts as uniform split. As shown in Table \ref{tab:ablation_study_result}, our algorithm with random split can only reach 2.4\% and our algorithm with uniform split can only reach 29.8\% of accuracy in MATH dataset, which are significantly lower than naive GRPO (61.20\%) and our algorithm with entropy based split (64.40\%), which shows the essence of entropy based split in our algorithm. We propose a possible reason to explain the the failure of random and uniform split strategies stems from the fact that these splits are not semantically-based: Within a single segment, there may exist both completely correct processes and completely incorrect processes. When such a segment receives a positive reward, the incorrect processes are incorrectly reinforced; conversely, when it receives a negative reward, the correct processes are incorrectly suppressed. Regardless of how the reward is assigned, a portion of the credit assignment will inevitably be problematic, ultimately leading to the collapse of model training.

We also did an ablation study on our predefined distribution for process-level advantage calculation. For comparison, we use the relative distribution of process-level advantage, which means the mean and standard deviation used to normalize and calculate process-level advantage is calculated each time based on the process rewards in the rollout. As shown in Table \ref{tab:ablation_study_result}, our algorithm with relative distribution for process-level advantage can reach 64.0\% of accuracy in MATH dataset and our algorithm with predefined distribution of process-level advantage can reach 64.4\% of accuracy in MATH dataset, where there is no significant difference. Nevertheless, we draw a graph of the change of accuracy in each epoch of our algorithm with relative distribution for process-level advantage and our algorithm with predefined distribution for process-level advantage in Figure \ref{fig:lineplot-trainepoch-GRPO} and Figure \ref{fig:ablation_study_result}, it can be seen that our algorithm with relative distribution for process-level advantage has its accuracy increase at first, but suddenly drop to nearly 0\% since the 6th epoch, which shows the great instability of relative distribution for process-level advantage calculation. We believe this is caused by the Premature Termination Problem we discussed in the previous section, when the model starts to reach a level of accuracy, some essential processes, which are still good but are relatively lower, can be an outlier and come with an extremely low advantage, finally leading to the collapse of the model. This ablation experiment proves the essence of predefined distribution for process-level advantage calculation in our algorithm, in order to both improve performance and stability of RL training. 

\section{Conclusion}

We present process Relative Policy Optimization, a critic-free method that aligns dense process rewards with sparse outcome rewards. PRPO normalizes PRM scores, aligns them with outcome advantages, and delivers stable per-token guidance. Experiments on mathematical reasoning benchmarks show consistent gains over GRPO and PRM-Avg while retaining critic-free efficiency.

\section{Limitations}

PRPO mitigates sparse rewards in GRPO by integrating dense PRM feedback, but it depends on PRM quality. In our work, we proposed a hypothesis about the possible collapse when training with process-only reward, and provide a possible mathematical proof of this hypothesis (shown in Appendix \ref{app:early_stop_proof}) which is incomplete because a lack of definition of the output distribution of PRMs. A more rigorous proof can be done to prove the point and provide more insights about how to avoid the collapse. The fixed prior mean and variance used for process-reward normalization assume PRM outputs lie in $[0,1]$, which is viable for most PRMs that normalize each output to $[0,1]$, but may not be applicable to PRMs with other ranges or distributions; extending normalization to PRMs with other ranges or distributions is an open direction. Future work will jointly optimize PRMs with the policy and explore adaptive segmentation that combines semantic cues.

\section{Potential Risks}

There are potential risks in using PRPO, which is the same as the risks in using GRPO. However, PRPO is more robust than GRPO because it has a more stable distribution of process-level advantage, which is more likely to converge to a stable distribution. For example, when the model starts to reach a level of accuracy, some essential processes, which are still good but are relatively lower, can be an outlier and come with an extremely low advantage that cannot be resolved by our distribution alignment mechanism, finally leading to the collapse of the model. This is a potential risk of using PRPO, and we believe it is a rare case and can be mitigated by using a more robust PRM or a more robust segmentation strategy. 

\bibliography{refs}
\onecolumn
\appendix
\section{Possible Mathematical Reason for Premature Collapse in Process-Only Reward \ref{prop:early_stop}}\label{app:early_stop_proof}
The gradient of the log-probability for the prefix is:
\[
\nabla_\theta \log p_\theta(x_{0:i}) = \sum_{j=0}^{i} \nabla_\theta \log \pi_\theta(x_j \mid x_{<j}).
\]
Under the policy gradient update derived from \eqref{eq:process_obj}, the expected change in $p_\theta(x_{0:t^\star})$ can be approximated by:
\[
\mathbb{E}\left[ \Delta p_\theta(x_{0:i}) \right] \approx \alpha \cdot \mathbb{E}\left[ \left( \sum_{j=0}^{i} A_j \nabla_\theta \log \pi_\theta(x_j \mid x_{\leq j}) \right)^\top \nabla_\theta \log p_\theta(x_{\leq j}) \right],
\]
where $\alpha$ is the learning rate. Assuming gradients for different $i$ are approximately orthogonal in expectation, we get:
\[
\mathbb{E}\left[ \Delta p_\theta(x_{0:i}) \right] \approx \alpha \cdot \mathbb{E}\left[ \sum_{j=0}^{i} A_j \| \nabla_\theta \log \pi_\theta(x_j \mid x_{\leq j}) \|^2 \right].
\]
Using the empirical averages from conditions (2) and (3):
\begin{align*}
\mathbb{E}\left[ \Delta p_\theta(x_{0:t^\star}) \right] 
&= \mathbb{E}\left[ \Delta (p_\theta(x_{0:t^\star-1})\cdot \pi_\theta(x_{t^\star} \mid x_{<t^\star})) \right] \\
&\approx \alpha \cdot \left( -a \sum_{i=0}^{t^\star-1} \mathbb{E}\left[ \| \nabla_\theta \log \pi_\theta(x_i \mid x_{<i}) \|^2 \right] + b \cdot \mathbb{E}\left[ \| \nabla_\theta \log \pi_\theta(x_{t^\star} \mid x_{<t^\star}) \|^2 \right] \right) \\
&\approx \alpha \cdot \left( -a \cdot (t^\star) + b \right) \cdot C,
\end{align*}
where $C > 0$ approximates the average squared gradient norm. When condition (4) holds ($a \cdot t^\star > b$), the term inside the parentheses is negative. Therefore:
\[
\mathbb{E}\left[ \Delta p_\theta(x_{0:t^\star}) \right] < 0.
\]
Consequently, with repeated updates, the probability $p_\theta(x_{0:t^\star})$ tends to decrease exponentially in practice. This makes the model increasingly unlikely to generate sequences reaching position $t^\star$, effectively encouraging earlier termination.

\twocolumn

\section{Entropy Based Segmentation Strategy}\label{app:entropy_based_segmentation}
\begin{algorithm}[H]\footnotesize
\caption{Entropy-based segmentation procedure.}
\label{alg:entropy_seg}
\begin{algorithmic}[1]
\Function{EntropySegmentation}{entropies, start\_idx, outLen, max\_branches=5, min\_gap=10}
    \If{$outLen - start\_idx < max\_branches + 1$}
        \State \Return $(start\_idx, outLen)$
    \EndIf

    \State anchors $\gets$ top-k entropy indices in entropies[start\_idx:outLen]
    \State anchors $\gets$ anchors + start\_idx
    \State anchors $\gets$ sort(anchors)

    \State filtered $\gets []$
    \For{anchor in anchors (left-to-right)}
        \If{distance(anchor, pre-kept) $\ge$ min\_gap}
            \State append anchor to filtered
        \EndIf
    \EndFor

    \State cuts $\gets []$; last\_cut $\gets$ start\_idx
    \For{$a$ in filtered}
        \If{$a - last\_cut \ge$ min\_gap}
            \State append $a$ to cuts; last\_cut $\gets a$
        \EndIf
    \EndFor

    \State segments $\gets []$; prev $\gets$ start\_idx
    \For{$c$ in cuts}
        \State append $(prev, c)$ to segments; prev $\gets c$
    \EndFor
    \State append $(prev, outLen)$ to segments

    \State sanitized $\gets []$; cur $\gets$ start\_idx
    \For{$(s, e)$ in segments}
        \State $s \gets \max(start\_idx, \min(outLen, s))$
        \State $e \gets \max(start\_idx, \min(outLen, e))$
        \If{$e \le s$}
            \State \textbf{continue}
        \EndIf
        \If{$s > cur$}
            \State append $(cur, s)$ to sanitized
        \EndIf
        \State append $(s, e)$ to sanitized; cur $\gets e$
    \EndFor
    \If{$cur < outLen$}
        \State append $(cur, outLen)$ to sanitized
    \EndIf

    \If{sanitized is empty}
        \State \Return $(start\_idx, outLen)$
    \Else
        \State \Return sanitized
    \EndIf
\EndFunction
\end{algorithmic}
\end{algorithm}
It's worth noting that we set k=5 and n=10 across all length of output sequences, except for one case when their is less than 5 tokens, we take the whole output sequence as one segment.

\section{Early Stop Point for Each Trainig}
\subsection{Qwen2.5-Math-1.5B}
\begin{itemize}
    \item GRPO: Epoch 2
    \item PRM-Avg: Epoch 7
    \item PURE: Epoch 10
    \item GRPO+\textbf{PRPO}: Epoch 8
    \item PRM-Avg+\textbf{PRPO}: Epoch 9
\end{itemize}
\subsection{Qwen2.5-Math-7B}
\begin{itemize}
    \item GRPO: Epoch 9
    \item PRM-Avg: Epoch 7
    \item PURE: Epoch 10
    \item GRPO+\textbf{PRPO}: Epoch 4
    \item PRM-Avg+\textbf{PRPO}: Epoch 3
\end{itemize}
\section{Disclosure of use of AI Tools}

We used the following AI tools for this research:
\begin{itemize}
\item Cursor for code revision and debugging. 
\item ChatGPT for evaluation of writing and proofreading. 
\item Gemini used to help design illustrations.
\end{itemize}

\section{Disclosure of license of the artifacts we used}
The artifacts we used are all open-source, and we used the following licenses:
\begin{itemize}
    \item The model Qwen2.5-Math-1.5B is released under the Apache License 2.0
    \item The model Qwen2.5-Math-7B is released under the Apache License 2.0
    \item The model Qwen2.5-Math-PRM-7B is released under the Apache License 2.0
    \item The dataset MATH is released under the MIT License
    \item The training framework VeRL is released under the Apache License 2.0
\end{itemize}
We strictly follow the license of the open-source artifacts we used. 

The data of AMC 2023, AIME 2025, and AIME 2024 are extracted from open website Art of Problem Solving \citep{aime_official_website}. We strictly follow the terms of use of this website and did not use the data obtained from the website for any commercial purpose. 

\end{document}